\documentclass[11pt]{article}

\usepackage[preprint]{acl}

\usepackage{subcaption}  
\usepackage{times}
\usepackage{latexsym}

\usepackage[T1]{fontenc}

\usepackage[utf8]{inputenc}

\usepackage{microtype}

\usepackage{inconsolata}

\usepackage{graphicx}

\usepackage{amsmath} 
\usepackage{enumitem}
\usepackage{array}
\usepackage{tabularx}
\usepackage{booktabs}
\usepackage{multirow}
\usepackage{bm} 
\usepackage[table]{xcolor}
\usepackage{colortbl}
\usepackage[breakable, skins, many]{tcolorbox}

\renewcommand{\paragraph}[1]{\smallskip\noindent\textbf{#1.}}
\renewcommand{\subparagraph}[1]{\smallskip\noindent\textbf{\underline{#1.}}}

\newcolumntype{L}[1]{>{\raggedright\arraybackslash}p{#1}}  

\newtcolorbox{promptbox}[1]{%
  enhanced,
  breakable,
  colback=gray!8,
  colframe=gray!40,
  colbacktitle=gray!70!black,
  coltitle=white,
  fonttitle=\bfseries,
  title=#1,
  boxrule=0.6pt,
  arc=2mm,
  left=3.5mm,right=3.5mm,top=2.5mm,bottom=2.5mm,
  fontupper=\small\ttfamily\raggedright\sloppy,
}

\title{When Abundance Conceals Weakness: Knowledge Conflict in Multilingual Models}

\author{
 \textbf{Jiaqi Zhao\textsuperscript{1}},
 \textbf{Qiang Huang\textsuperscript{1}\thanks{Qiang Huang is the corresponding author.}},
 \textbf{Haodong Chen\textsuperscript{1}},
 \textbf{Xiaoxing You\textsuperscript{2}},
 \textbf{Jun Yu\textsuperscript{1}}
\\
 \textsuperscript{1}Harbin Institute of Technology (Shenzhen),~\textsuperscript{2}Hangzhou Dianzi University
\\
   \{zhaojiaqi, chen.haodong\}@stu.hit.edu.cn,~\{huangqiang, yujun\}@hit.edu.cn, \\ youxiaoxing@hdu.edu.cn
}

\begin{document}

\maketitle
\begin{abstract}
Large Language Models (LLMs) encode vast world knowledge across multiple languages, yet their internal beliefs are often unevenly distributed across linguistic spaces. 
When external evidence contradicts these language-dependent memories, models encounter \emph{cross-lingual knowledge conflict}, a phenomenon largely unexplored beyond English-centric settings.
We introduce \textbf{CLEAR}, a \textbf{C}ross-\textbf{L}ingual knowl\textbf{E}dge conflict ev\textbf{A}luation f\textbf{R}amework that systematically examines how multilingual LLMs reconcile conflicting internal beliefs and multilingual external evidence. 
CLEAR decomposes conflict resolution into four progressive scenarios, from multilingual parametric elicitation to competitive multi-source cross-lingual induction, and systematically evaluates model behavior across two complementary QA benchmarks with distinct task characteristics.
We construct multilingual versions of ConflictQA and ConflictingQA covering 10 typologically diverse languages and evaluate six representative LLMs.
Our experiments reveal a task-dependent decision dichotomy. 
In reasoning-intensive tasks, conflict resolution is dominated by language resource abundance, with high-resource languages exerting stronger persuasive power. 
In contrast, for entity-centric factual conflicts, linguistic affinity, not resource scale, becomes decisive, allowing low-resource but linguistically aligned languages to outperform distant high-resource ones. 
\end{abstract}

\section{Introduction}
\label{sec:intro}

Large Language Models (LLMs) are trained on vast corpora and encode substantial world knowledge in their parameters \cite{hurst2024gpt, team2023gemini, yang2025qwen3, grattafiori2024llama}. 
In practice, however, modern LLM systems rarely rely on parametric memory alone. 
To mitigate errors and hallucinations, external information is commonly injected at inference time--most prominently via Retrieval-Augmented Generation (RAG) \cite{chen2022rich, cattan2025dragged, park2025investigating}, where retrieved documents provide additional evidence for answering user queries.

\begin{figure}[t]
  \centering
  \includegraphics[width=0.99\columnwidth]{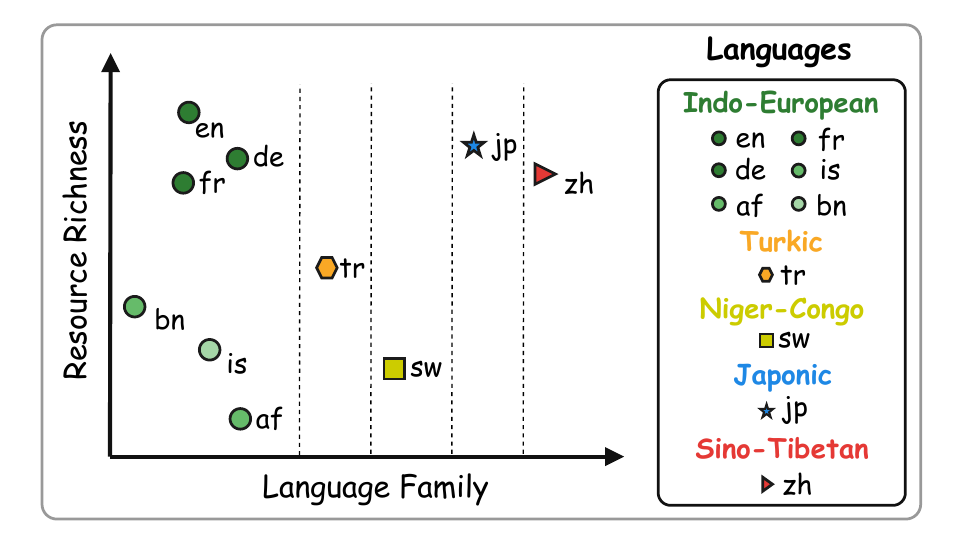}
  \vspace{-1.0em}
  \caption{language distribution in the CLEAR framework: languages are mapped based on their resource richness and taxonomic family, enabling a systematic study of how linguistic affinity and data scale influence cross-lingual knowledge conflict resolution.}
  \label{fig:language-distribution}
  \vspace{-0.5em}
\end{figure}

Yet, introducing external evidence gives rise to a critical failure mode: \textbf{knowledge conflict} \cite{chen2022rich}, in which a model's internally stored belief contradicts information presented in context. 
How LLMs react to and resolve such conflicts is central to system robustness and faithfulness, particularly in high-stakes or evidence-grounded applications.
Crucially, these systems are increasingly \textbf{multilingual}. 
In real-world RAG pipelines, queries and retrieved sources often span multiple languages, requiring models to reconcile evidence across linguistic boundaries while producing coherent outputs.
Despite this reality, existing studies remain largely English-centric, leaving the mechanisms underlying \textbf{Cross-Lingual Knowledge Conflict (CLKC)} largely unexplored.
The challenge of CLKC is threefold.
\begin{itemize}[nolistsep,left=12pt]
  \item \textbf{Multilingual LLMs exhibit language-conditioned memories}: knowledge correct in one language may be incomplete or incorrect in another \cite{kassner2021multilingual}. 
  Prior multilingual studies focus on source-language preference \cite{park2025investigating}, implicitly assuming fixed internal beliefs, leaving unclear how query language activates parametric memory and when multilingual evidence overrides or reinforces it.

  \item \textbf{Existing alignment studies often conflate distinct knowledge types.} 
  While cross-lingual consistency is typically measured at the output level \cite{wang-etal-2025-linguistic}, \emph{entity-centric factual knowledge} differs fundamentally from that requiring \emph{multi-step logical reasoning}, and their interaction with multilingual evidence remains underexplored.

  \item \textbf{Research on knowledge conflict has been overwhelmingly monolingual} \cite{longpre2021entity, chen2022rich, xie2024adaptive, jin2024tug}, particularly centering on English-centric scenarios. 
  Consequently, it remains unclear how conflict resolution operates when queries, evidence, and prior knowledge reside in different linguistic spaces.
\end{itemize}

To address these gaps, we propose \textbf{CLEAR}, a \textbf{C}ross-\textbf{L}ingual knowl\textbf{E}dge conflict ev\textbf{A}luation f\textbf{R}amework for systematically studying how multilingual LLMs navigate knowledge conflicts.
As illustrated in Figure \ref{fig:language-distribution}, CLEAR spans 10 languages chosen to vary in both \emph{resource abundance} and \emph{linguistic affinity}, enabling controlled analysis of whether cross-lingual decision-making is driven primarily by training data scale or structural proximity between languages.
Unlike prior work that treats knowledge conflict as binary, CLEAR decomposes conflict resolution into four progressive tasks, ranging from multilingual parametric elicitation to competitive multi-source cross-lingual induction, and evaluates model behavior on two QA benchmarks with different conflict patterns.

We conduct extensive experiments on two newly curated multilingual benchmarks: ConflictQA-PopQA and ConflictQA-StrategyQA, covering 10 typologically diverse languages. 
Our results reveal a task-dependent decision dichotomy: in reasoning-intensive tasks, conflict resolution is dominated by language resource abundance, whereas in entity-centric factual tasks, linguistic affinity, rather than data scale, emerges as the primary driver of persuasion. 
Notably, low-resource but linguistically aligned languages can exert stronger influence than distant high-resource languages, exposing an abundance-weakness paradox in multilingual LLMs.
The contributions of this work are summarized as follows:
\begin{itemize}[nolistsep,left=12pt]
  \item \textbf{Cross-Lingual Knowledge Conflict (CLKC).}
  We introduce CLKC as a new evaluation paradigm that reframes knowledge conflict as a tension between \emph{language-dependent parametric beliefs} and \emph{multilingual external evidence}, enabling systematic analysis of belief activation and revision across languages.

  \item \textbf{Task-Dependent Cross-Lingual Conflict Resolution.}
  We uncover a novel behavioral pattern in CLKC: conflict resolution in reasoning-intensive tasks is driven primarily by \emph{language resource abundance}, whereas entity-centric factual conflicts are governed by \emph{linguistic affinity} rather than data scale.

  \item \textbf{Multilingual Conflict Benchmarks.}
  We construct multilingual versions of \emph{ConflictQA-PopQA} and \emph{ConflictQA-StrategyQA} across 10 typologically diverse languages, enabling controlled comparison of cross-lingual knowledge conflict in entity-centric and reasoning-intensive QA settings.
\end{itemize}

\section{Related Work}
\label{sec:related_work}

Our work lies at the intersection of Retrieval-Augmented Generation (RAG), knowledge conflict in LLMs, and cross-lingual consistency, aiming to understand how multilingual LLMs arbitrate between language-dependent parametric knowledge and multilingual external evidence.

\paragraph{Language-Aware RAG}
RAG is widely adopted to mitigate hallucinations, yet it introduces conflicts when retrieved sources disagree \citep{chen2022rich, cattan2025dragged}. 
Although prior work studies language preference in multilingual RAG \citep{park2025investigating}, the mechanisms of cross-lingual multi-source competition remain unclear. 
Extending the \emph{tug-of-war} view \citep{jin2024tug}, we model this interaction as a \textit{Quadratic Knowledge Nexus}, disentangling query-language bias from source-language authority. 
This formulation exposes two key drivers of conflict resolution: \textit{Language Dominance} and the \textit{Query Priming Effect}.

\begin{figure*}[t]
  \centering
  \includegraphics[width=0.99\textwidth]{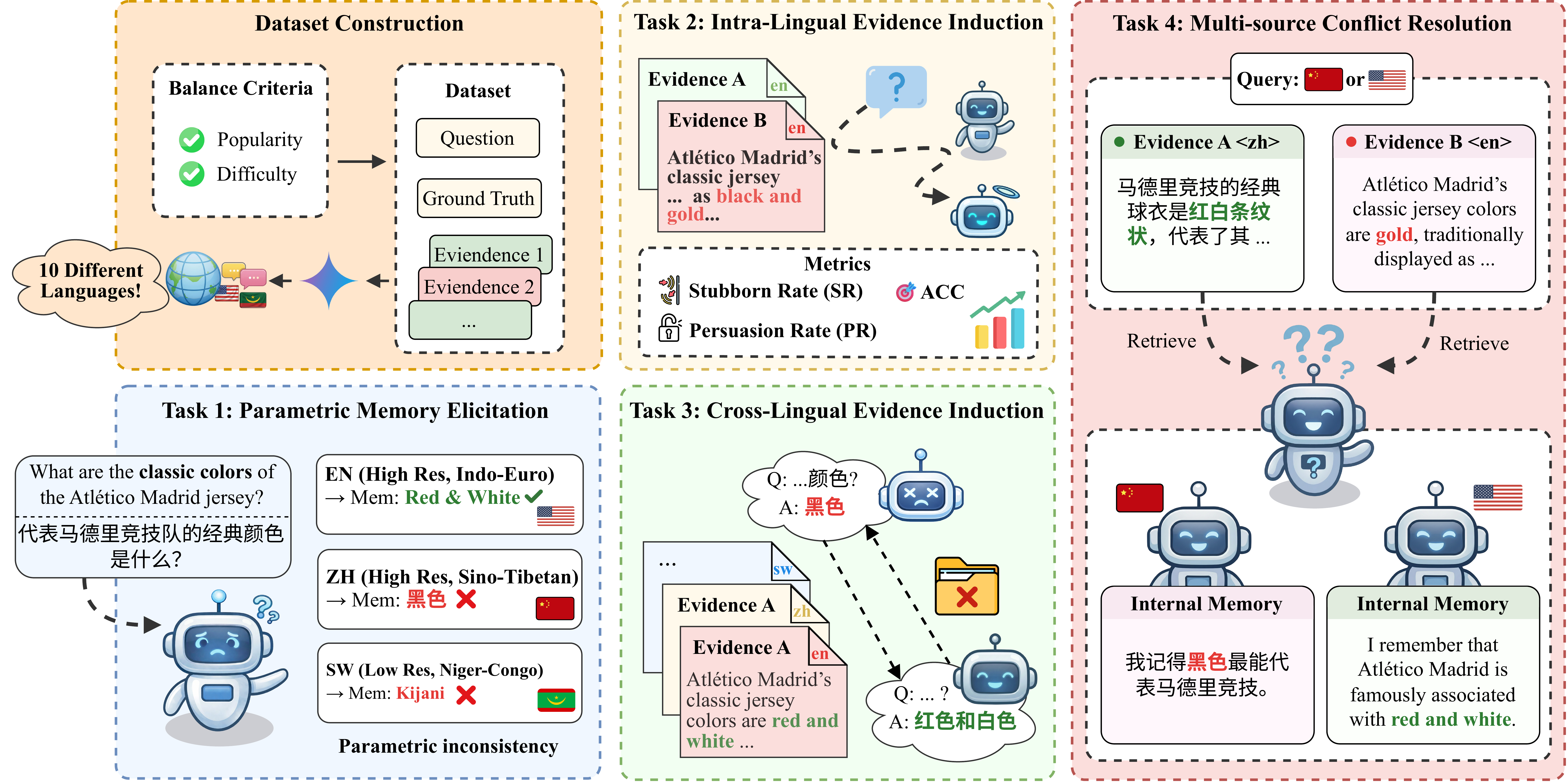}
  \caption{Overview of the \textbf{CLEAR} framework.}
  \label{fig:lang_family}
  \vspace{-0.5em}
\end{figure*}

\paragraph{Knowledge Conflict in LLMs} 
Knowledge conflict arises when external evidence contradicts an LLM's parametric memory. 
Prior studies characterize such behavior as adaptive or resistant \citep{xie2024adaptive}, analyze its disruptive effects \citep{sun2025seen}, or regulate reliance on internal versus external knowledge \citep{bi2025parameters, wang2025adacad}. 
However, these efforts are largely monolingual in English only. Even work on evidence convincingness \citep{wan2024evidence, tan2024blinded} typically ignores language. 
We show instead that knowledge conflict is fundamentally \emph{language-conditioned}, shaped by how knowledge is encoded across linguistic spaces.

\paragraph{Cross-Lingual Knowledge Consistency}
Multilingual LLMs often encode the same entity inconsistently across languages, a phenomenon known as \textit{cross-lingual asymmetry} \citep{kassner2021multilingual, ai2025knowledge}. 
Existing work focuses on linguistic connectivity or translation-level consistency in closed-book settings \citep{wang-etal-2025-linguistic, mitrovic2025llms}.
We move beyond static evaluation to study \emph{dynamic interaction}, using these asymmetries to probe how language-specific memories compete when external evidence is introduced. 


\section{The CLEAR Framework}
\label{sec:framework}

We present \textbf{CLEAR}, a framework for eliciting multilingual parametric memory from LLMs, constructing controlled counter-memory across languages, and evaluating how models resolve cross-lingual knowledge conflicts. 
Our design ensures that all conflicts are grounded in the model's own internal beliefs rather than annotation artifacts.

\subsection{Multilingual Dataset Construction}
\label{sec:framework:construct}

\paragraph{Question Answering (QA) Datasets}
Following prior work \cite{longpre2021entity, chen2022rich}, we adopt the QA task as the primary testbed.
We employ a QA benchmark suite based on \textbf{ConflictQA} \cite{xie2024adaptive}, which is derived from PopQA \citep{mallen2023not} (entity-centric factual knowledge) and StrategyQA \citep{geva2021did} (commonsense questions with higher reasoning demands).

From ConflictQA, we curate high-quality subsets by filtering for knowledge popularity, question type, and difficulty:
\begin{itemize}[nolistsep,left=12pt]
  \item \textbf{ConflictQA-PopQA} includes 898 entity-centric queries with Wikipedia-based evidence, refined via human annotation to ensure strong entity-level contradictions.

  \item \textbf{ConflictQA-StrategyQA} contains 1,000 reasoning-intensive samples, where external evidence is synthesized by LLMs to preserve fluency while minimizing interference with implicit reasoning.
\end{itemize}

\paragraph{Multilingual Dataset Construction}
To study cross-lingual dynamics, we translate all benchmarks into 10 diverse languages:
\begin{displaymath}
  \mathcal{L} = \{\text{af, bn, de, fr, is, ja, sw, zh, tr, en}\}, 
\end{displaymath}
covering five language families and a wide range of resource levels \citep{joshi-etal-2020-state}.
Translation is performed using Gemini-2.5-Pro \cite{team2023gemini}, followed by human verification to ensure semantic fidelity.

For entity-centric tasks, we apply an \emph{entity-presence constraint}: the ground-truth entity must appear explicitly in supportive evidence and be absent or replaced in conflicting evidence. This guarantees that model decisions reflect a genuine choice between \emph{parametric memory} and \emph{external cues}, rather than surface ambiguity.

\subsection{Tasks}
\label{sec:framework:tasks}

CLEAR decomposes cross-lingual knowledge conflict into four progressively complex tasks, each isolating interactions between language-dependent parametric memory and external evidence.

\paragraph{Task 1: Parametric Memory Elicitation}
In a closed-book setting without external context, LLMs answer semantically equivalent queries posed in different languages, relying solely on their internal parametric memory. 
This task directly probes how factual knowledge is unevenly encoded across linguistic spaces.

Unlike prior work that studies conflicts within a single (typically English) language, this task exposes \textbf{cross-lingual parametric asymmetry}, cases where the same model holds correct beliefs in one language but incorrect or missing beliefs in another.
We retain incorrect answers, as they reflect biased memories genuinely stored in model parameters and form the basis for subsequent induction tasks.

\paragraph{Task 2: Intra-Lingual Evidence Induction}
To assess the robustness of parametric memory within a single language, we pair each query with \textbf{one contradictory evidence snippet in the same language}. 
This creates an intra-lingual conflict between internal memory and external context.

This task measures whether models persist in their parametric belief (\textbf{stubbornness}) or revise it to follow evidence (\textbf{persuasion}).
By stratifying queries according to entity popularity, we further analyze how the strength of memorized knowledge affects susceptibility to induction.
Table~\ref{tab:SR_PR} reports the resulting Stubborn Rate (SR) and Persuasion Rate (PR) across 10 languages on two datasets.

\paragraph{Task 3: Cross-Lingual Evidence Induction}
Building on Task 2, we replace same-language evidence with evidence written in a different language. 
Specifically, queries are presented in the target language $L_{tgt}$, while supportive or conflicting evidence is provided in a source language $L_{src}$ ($L_{src} \neq L_{tgt}$); answers must be produced in $L_{tgt}$.

This task isolates the \textbf{cross-lingual persuasive power} of evidence, examining whether models can revise beliefs across linguistic boundaries and how this ability varies when parametric memory is initially correct versus incorrect.

\paragraph{Task 4: Multi-Source Conflict Resolution}
To simulate realistic, high-pressure settings, we present the model with \textbf{two explicitly contradictory evidence sources} expressed in different languages. 
Using a symmetric 2$\times$2 permutation design for each language pair, we control for query-language bias and isolate source competition.

This task investigates the dynamic interplay between parametric memory and multilingual evidence competition, focusing on two states:
\begin{enumerate}[nolistsep,label*=(\arabic*),left=12pt]
  \item \textbf{Memory-Supportive Competition}, where one source aligns with internal belief,
  
  \item \textbf{Memory-Conflicting Competition}, where internal belief is incorrect or absent.
\end{enumerate}

By analyzing outcomes across language pairs, we test whether certain languages exhibit inherent dominance that can override both memory and competing sources.

\subsection{Evaluation Metrics}
\label{sec:framework:metrics}

We define three metrics to quantify the frequency of LLMs adhering to their parametric memory or being influenced by external evidence. 
Let $a_{out} = M(q^{L_{query}}, \mathcal{C})$ denote the model output given query $q$ and evidence set $\mathcal{C} = \{C_i^{L_i}\}_{i=1}^k$, where $k \geq 0$, and $L_query$ and $L_i$ represent the language of the query and the $i$-th evidence. When $\mathcal{C} = \emptyset$, the process reduces to pure parametric elicitation.

\paragraph{Stubborn Rate (SR)} 
The probability that the model preserves the correct parametric answer $y$ when confronted with conflicting evidence $C^L_{con}$:
\begin{displaymath}
  SR = P(a_{out} = y \mid a^L_{param} = y, \mathcal{C} = \{C^L_{con}\}).
\end{displaymath}
This metric measures the model's resilience against misleading information when its internal memory is correct. 
High $SR$ indicates that the model's parametric memory is robust and difficult to override by linguistic manipulation.

\paragraph{Persuasion Rate (PR)} 
The probability of correcting an incorrect parametric belief $a^L_{param}$ ($a^L_{param} \neq y$) given supportive evidence $C^L_{sup}$:
\begin{displaymath}
  PR = P(a_{out} = y \mid a^L_{param} \neq y, \mathcal{C} = \{C^L_{sup}\}).
\end{displaymath}
This metric assesses the model's ability to rectify its internal errors when provided with correct external evidence.
For multi-choice or binary tasks like StrategyQA, $SR$ and $PR$ are complementary ($SR \approx 1 - PR_{induced}$), whereas for entity-centric tasks like PopQA, they provide independent insights into model behavior. 

\paragraph{Accuracy (ACC)}
We define accuracy as the probability of producing the ground-truth answer under conflict, $ACC = P(a_{out}=y)$, where $y$ is the ground truth. 
ACC summarizes overall reliability when reconciling parametric memory with cross-lingual evidence contexts $\mathcal{C}$.

\section{Experiments}
\label{sec:expt}

\begin{figure*}[t]
  \centering
  \includegraphics[width=0.99\textwidth]{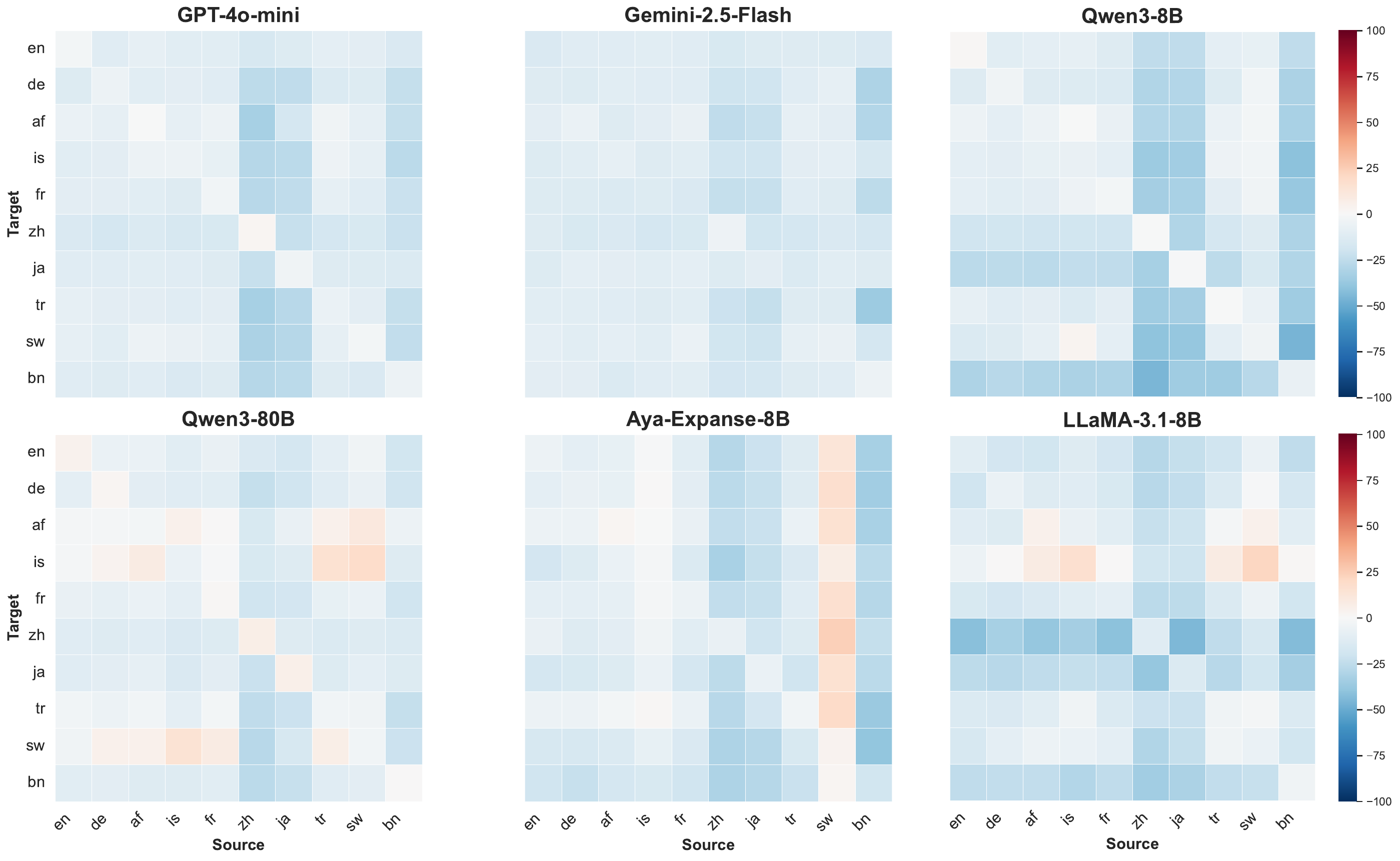}
  \vspace{-0.5em}
  \caption{\textbf{Task-dependent Persuasion Rate difference.} $\Delta$ denotes the difference in Persuasion Rate between PopQA and StrategyQA (PopQA $-$ StrategyQA), highlighting task-specific induction behavior.}
  \label{fig:task3_heatmap}
  \vspace{-0.5em}
\end{figure*}

\subsection{Experimental Setup}
\label{sec:expt:setup}

To investigate cross-lingual knowledge conflict mechanisms, we evaluate our framework across a diverse set of Large Language Models (LLMs), encompassing both proprietary and open-source systems. For proprietary models, we include \textbf{GPT-4o-mini} \cite{hurst2024gpt} and \textbf{Gemini-2.5-Flash}\cite{team2023gemini}, both accessed through the \textbf{OpenRouter API} to ensure consistent response generation. For open-source models, we select representative systems spanning various parameter scales and multilingual optimizations: \textbf{Qwen3-8B}, \textbf{Qwen3-80B} \cite{yang2025qwen3}, \textbf{Llama-3.1-8B} \cite{grattafiori2024llama}, and \textbf{Aya-Expanse-8B} \cite{ustun2024aya}. 

All open-source models are deployed and executed on a single \textbf{NVIDIA A100 GPU (80GB)}. Notably, for models equipped with advanced reasoning or ``thinking'' capabilities, such as Qwen3-8B, we explicitly \textbf{disable the thinking process} during evaluation. This ensures that the model's responses reflect its direct parametric knowledge and its immediate reaction to linguistic conflicts, rather than an iterative self-correction process that could mask the underlying cross-lingual misalignment. 

To robustly evaluate multilingual outputs under minor surface variations (e.g., spelling and name formatting), we adopt an \textbf{LLM-as-a-judge} protocol. 
We use \textbf{Gemini-2.5-Flash} as the judge to determine whether the model output $a_{out}$ matches the language-specific ground truth $y^{L}$ for each query. Table~\ref{tab:closed_book_performance} summarizes closed-book answer accuracy for each model across languages.
Additional implementation details can be found in Appendix \ref{app:details}.

\begin{table}[t]
\centering
\small
\setlength{\tabcolsep}{3pt}
\renewcommand{\arraystretch}{1.5}
\resizebox{\columnwidth}{!}{%
\begin{tabular}{lcccccccccc}
\toprule
\textbf{Models} & \textbf{af} & \textbf{bn} & \textbf{de} & \textbf{en} & \textbf{fr} & \textbf{is} & \textbf{ja} & \textbf{sw} & \textbf{tr} & \textbf{zh} \\
\midrule
\rowcolor[gray]{0.92}
\multicolumn{11}{c}{\textbf{PopQA}} \\
\midrule
\textbf{GPT-4o-mini}        & 47.3 & 26.1 & 51.4 & 52.7 & 49.7 & 41.0 & 34.4 & 40.5 & 43.8 & 27.6 \\
\textbf{Gemini-2.5-Flash}   & 54.1 & 42.2 & 58.7 & 59.9 & 56.3 & 53.9 & 45.2 & 49.8 & 51.4 & 40.1 \\
\cmidrule(lr){1-11}
\textbf{Llama-3.1-8B}       & 34.0 & 15.4 & 39.4 & 43.5 & 36.6 & 23.9 & 18.7 & 26.5 & 29.4 & 19.2 \\
\textbf{Qwen3-8B}           & 23.2 & 10.6 & 29.1 & 32.6 & 30.2 & 14.0 & 17.8 & 14.8 & 21.7 & 23.9 \\
\textbf{Qwen3-80B}          & 45.7 & 22.0 & 47.8 & 53.8 & 47.6 & 35.0 & 27.4 & 36.0 & 37.8 & 34.6 \\
\textbf{Aya-Expanse-8B}     & 30.0 & 10.1 & 35.3 & 38.4 & 35.6 & 16.3 & 22.3 & 18.6 & 32.2 & 23.9 \\
\midrule
\rowcolor[gray]{0.92}
\multicolumn{11}{c}{\textbf{StrategyQA}} \\
\midrule
\textbf{GPT-4o-mini}        & 68.1 & 68.2 & 70.8 & 74.1 & 71.6 & 65.8 & 70.4 & 68.7 & 68.1 & 70.0 \\
\textbf{Gemini-2.5-Flash}   & 64.7 & 62.6 & 67.7 & 71.1 & 65.2 & 62.3 & 64.9 & 64.4 & 64.3 & 64.2 \\
\cmidrule(lr){1-11}
\textbf{Llama-3.1-8B}       & 51.4 & 54.4 & 54.9 & 65.6 & 58.3 & 47.0 & 55.5 & 50.8 & 54.3 & 61.7 \\ 
\textbf{Qwen3-8B}           & 53.2 & 52.4 & 57.7 & 65.5 & 60.4 & 47.7 & 54.0 & 48.0 & 52.6 & 57.5 \\
\textbf{Qwen3-80B}          & 55.7 & 59.5 & 65.0 & 70.2 & 67.3 & 53.4 & 62.6 & 52.9 & 59.8 & 63.5 \\
\textbf{Aya-Expanse-8B}     & 49.6 & 48.0 & 55.0 & 62.8 & 57.9 & 50.0 & 57.4 & 49.6 & 52.7 & 54.0 \\
\bottomrule
\end{tabular}%
}
\caption{\textbf{Closed-book accuracy on ConflictQA-PopQA (Step 1 in Figure 1).} We report response accuracy for eight LLMs, including two closed-source and four open-source models.}
\label{tab:closed_book_performance}
\end{table}

\begin{table*}[t]
\centering
\setlength{\tabcolsep}{5pt}
\renewcommand{\arraystretch}{1.15}
\resizebox{\textwidth}{!}{%
\begin{tabular}{l cccccccccc cccccccccc}
\toprule
\multirow{2.5}{*}{\textbf{Model}}
& \multicolumn{10}{c}{\textbf{Stubborn Rate}} 
& \multicolumn{10}{c}{\textbf{Persuasion Rate}} \\
\cmidrule(lr){2-11}\cmidrule(lr){12-21}
& af & bn & de & en & fr & is & ja & sw & tr & zh
& af & bn & de & en & fr & is & ja & sw & tr & zh \\
\midrule

\rowcolor[gray]{0.92}
\multicolumn{21}{c}{\textbf{PopQA}} \\
\midrule

\textbf{GPT-4o-mini}
& 9.2 & 13.3 & 9.7 & 10.2 & 9.4 & 10.9 & 11.3 & 11.0 & 11.7 & 13.3
& 85.6 & 81.2 & 78.9 & 79.5 & 79.2 & 81.3 & 80.5 & 82.2 & 80.0 & 84.5 \\

\textbf{Gemini-2.5-Flash}
& 10.1 & 10.6 & 8.7 & 10.2 & 8.7 & 10.3 & 10.8 & 10.7 & 10.2 & 12.8
& 79.1 & 83.8 & 79.0 & 76.1 & 77.0 & 80.2 & 81.1 & 82.9 & 78.7 & 84.0 \\

\cmidrule(lr){1-21}

\textbf{LLaMA-3.1-8B}
& 15.4 & 13.0 & 16.7 & 13.3 & 14.3 & 26.1 & 19.6 & 12.2 & 17.4 & 21.5
& 81.6 & 78.6 & 80.2 & 79.1 & 77.9 & 77.8 & 78.4 & 75.2 & 79.7 & 77.6 \\

\textbf{Qwen3-8B}
& 9.6 & 15.8 & 16.5 & 17.1 & 15.9 & 13.5 & 20.6 & 10.5 & 12.3 & 18.6
& 84.2 & 78.8 & 83.4 & 87.6 & 83.3 & 78.6 & 84.2 & 75.8 & 83.6 & 83.9 \\

\textbf{Qwen3-80B}
& 11.7 & 13.1 & 16.1 & 17.8 & 14.8 & 12.1 & 14.2 & 9.6 & 15.0 & 15.4
& 82.0 & 80.6 & 78.0 & 78.3 & 78.6 & 80.3 & 84.2 & 79.7 & 80.3 & 82.3 \\

\textbf{Aya-Expanse-8B}
& 8.2 & 8.8 & 12.6 & 13.3 & 13.8 & 13.0 & 15.0 & 12.6 & 12.8 & 16.7
& 86.8 & 67.9 & 86.2 & 87.3 & 87.0 & 77.3 & 86.1 & 70.3 & 86.4 & 85.5 \\

\cmidrule(lr){1-21}

\textit{Average}
& 10.7 & 12.4 & 13.4 & 13.7 & 12.8 & 14.3 & 15.3 & 11.1 & 13.2 & 16.4
& 83.2 & 78.5 & 81.0 & 81.3 & 80.5 & 79.3 & 82.4 & 77.7 & 81.5 & 83.0 \\

\midrule

\rowcolor[gray]{0.92}
\multicolumn{21}{c}{\textbf{StrategyQA}} \\
\midrule

\textbf{GPT-4o-mini}
& 34.7 & 32.8 & 37.0 & 39.3 & 37.6 & 34.8 & 32.8 & 34.4 & 32.2 & 34.3
& 86.2 & 86.8 & 84.6 & 82.2 & 82.8 & 87.4 & 84.5 & 84.7 & 86.8 & 82.3 \\

\textbf{Gemini-2.5-Flash}
& 17.8 & 19.3 & 20.8 & 21.0 & 19.5 & 19.1 & 19.3 & 17.6 & 17.6 & 21.0
& 92.1 & 89.8 & 92.6 & 91.7 & 92.5 & 93.4 & 91.7 & 90.7 & 91.3 & 89.4 \\

\cmidrule(lr){1-21}

\textbf{LLaMA-3.1-8B}
& 32.7 & 19.5 & 24.6 & 26.1 & 24.4 & 47.2 & 13.5 & 24.6 & 23.4 & 17.3
& 76.5 & 83.1 & 87.1 & 90.1 & 88.0 & 61.5 & 93.0 & 81.7 & 84.3 & 89.6 \\

\textbf{Qwen3-8B}
& 25.2 & 23.7 & 32.9 & 34.4 & 34.3 & 22.0 & 30.2 & 26.9 & 24.3 & 34.8
& 89.7 & 86.3 & 87.7 & 86.1 & 85.9 & 86.4 & 85.2 & 80.2 & 84.2 & 84.0 \\

\textbf{Qwen3-80B}
& 50.6 & 44.4 & 53.2 & 51.4 & 51.0 & 37.8 & 48.2 & 33.7 & 40.5 & 49.6
& 84.9 & 80.5 & 75.7 & 73.8 & 77.1 & 86.9 & 78.6 & 82.8 & 83.6 & 75.6 \\

\textbf{Aya-Expanse-8B}
& 17.1 & 15.4 & 13.5 & 11.8 & 12.8 & 22.0 & 11.9 & 31.9 & 12.7 & 12.8
& 84.9 & 86.7 & 93.3 & 93.6 & 92.6 & 79.2 & 92.5 & 66.5 & 90.3 & 93.3 \\

\cmidrule(lr){1-21}

\textit{Average}
& 29.7 & 25.9 & 30.3 & 30.7 & 29.9 & 30.5 & 26.0 & 28.2 & 25.1 & 28.3
& 85.7 & 85.5 & 86.8 & 86.3 & 86.5 & 82.5 & 87.6 & 81.1 & 86.8 & 85.7 \\

\bottomrule
\end{tabular}%
}
\caption{\textbf{Stubborn Rate and Persuasion Rate across 10 languages on PopQA and StrategyQA.} Higher SR indicates stronger reliance on parametric memory, while higher PR reflects more effective evidence-based correction.}
\label{tab:SR_PR}
\end{table*}

\subsection{Dataset Disparities: Facts vs. Reasoning}
\label{sec:expt:dataset-disparities}

Our results in Table \ref{tab:SR_PR} reveal a significant disparity in how LLMs reconcile conflicts across different knowledge types. In the entity-centric \textbf{PopQA}, models exhibit a relatively low average Stubborn Rate (SR) of 13.4\%, coupled with a high Persuasion Rate (PR) of 81.0\%. This suggests that for simple factual entities, LLMs are highly susceptible to external evidence, even when it directly contradicts their internal parametric beliefs. 

In contrast, the reasoning-heavy \textbf{StrategyQA} dataset elicits a much stronger internal resistance, with an average SR of 28.4\%, nearly double that of PopQA. This indicates that when knowledge is embedded within a multi-step logical chain, models are significantly more prone to ignore external evidence in favor of their internal reasoning pathways. Interestingly, PR remains high across both datasets (averaging over 80\%), demonstrating that models are generally capable of utilizing correct external context to rectify incorrect memories, regardless of the task complexity.

\subsection{Linguistic Variances and Cross-Lingual Stability}
\label{sec:expt:linguistic-variances}

Linguistic factors play a crucial role in conflict resolution dynamics. 
As shown in Table \ref{tab:SR_PR}, \textbf{PopQA} exhibits clear variation across languages. \textbf{Chinese (zh)} and \textbf{Japanese (ja)} achieve the highest SR (16.4\% and 15.3\% respectively), suggesting stronger alignment or confidence in entity-centric parametric representations within these scripts.
In contrast, lower-resource languages like \textbf{Afrikaans (af)} display the lowest SR (10.7\%), indicating greater reliance on external evidence, likely due to weaker or less stable internal representations.

A different pattern emerges in \textbf{StrategyQA}, where stubbornness aligns more closely with language resource abundance.
High-resource languages such as \textbf{English (en)} and \textbf{German (de)} exhibit substantially higher SR values (averaging above 30\%). In contrast, lower-resource languages, including \textbf{Turkish (tr)} and \textbf{Bengali (bn)}, are more readily persuaded.
This contrast suggests that model confidence in multi-step reasoning is strongly conditioned by the availability of training data in a given language: resource-rich languages tend to support more resilient (and potentially overconfident) internal reasoning chains.

\subsection{The Dual-Pathway of Cross-Lingual Authority}
\label{sec:expt:dual-pathway}

Figure~\ref{fig:task3_heatmap} reports the gap in Persuasion Rate (PR) between PopQA and StrategyQA across six models. 
A consistent and striking pattern emerges: across all models, high-resource non-Latin languages such as \textbf{Chinese (zh)}, \textbf{Japanese (ja)}, and \textbf{Bengali (bn)} exhibit substantially lower PR on PopQA than on StrategyQA, as reflected by the deep blue regions in the heatmap.
This disparity is especially pronounced in smaller open-source models, such as \textbf{LLaMA-3.1-8B} and \textbf{Aya-Expanse-8B}, where these languages demonstrate strong authority in logical reasoning yet fail to effectively correct entity-centric factual errors.

In contrast, low-resource languages including \textbf{Icelandic (is)} and \textbf{Swahili (sw)} display the opposite trend. 
Their PR on PopQA is comparable to, and in some cases exceeds, their performance on StrategyQA. Notably, in \textbf{Aya-Expanse-8B}, these languages yield neutral or slightly positive PR gaps (warm colors), indicating greater effectiveness in resolving entity-centric conflicts than their higher-resource counterparts.
Together, these results reveal an unexpected inversion in cross-lingual behavior, suggesting that knowledge authority in LLMs operates along two distinct pathways:

\paragraph{Path 1: The Logic-Resource Path} 
In reasoning tasks, the persuasive power of a language is almost strictly linear to its pre-training volume. 
High-resource languages like \textbf{Chinese (zh), Japanese (ja), and Bengali (bn)} emerge as dominant anchors. Their abundance of training data provides a dense and robust logical scaffold. 
When these languages serve as the source of evidence, they can effectively \emph{rescue} models from reasoning fallacies in any target language. In this domain, the model's strength is a direct artifact of scale: more data yields a more resilient logic engine.

\paragraph{Path 2: The Representation-Affinity Path} 
However, this resource-driven authority crumbles when the task shifts to entity-centric factual conflicts (PopQA). 
Despite their logical prowess, \textbf{zh, ja, and bn} show a surprising inability to correct factual errors involving Latin-based entities. Here, the \textbf{Script Barrier} acts as a profound isolator; the symbolic distance between non-Latin scripts and the original entity names (often stored in a Latin-centric global latent space) hinders effective knowledge retrieval and alignment.

In a remarkable reversal, low-resource languages that share the Latin script, such as \textbf{Swahili (sw), Icelandic (is), and Afrikaans (af)}, become the most effective rescuers in PopQA. 
Their strength lies not in the volume of what they know, but in the \emph{proximity} of how they represent it. Because they share a common alphabet and morphological roots with the target entities, they provide a more direct and reliable address in the model's memory, bypassing the alignment rot that plagues even the most data-rich non-Latin languages.

\subsection{Unveiling the ``Abundance-Weakness'' Paradox}
\label{sec:expt:abundance-weakness}

Taken together, these results reveal an abundance-weakness paradox: data abundance in one cognitive dimension can conceal structural fragility in another.
As depicted in Figure~\ref{fig:task3_heatmap}, high-resource non-Latin languages show strong authority in reasoning-intensive tasks but consistently underperform in correcting entity-centric factual conflicts, indicating that their apparent multilingual competence is driven by logical reasoning rather than reliable cross-lingual factual alignment.
This disparity suggests a \textbf{structurally fragmented cross-lingual latent space}, where fluent reasoning and factual grounding diverge across scripts, creating fertile conditions for cross-lingual knowledge conflict.

\begin{table*}[t]
\small
\centering
\renewcommand{\arraystretch}{1.2}
\begin{tabular}{llcccccccccc}
\toprule
\multirow{2.5}{*}{\textbf{Dataset}} & \multirow{2.5}{*}{\textbf{Model}}
& \multicolumn{5}{c}{\textbf{Q = Pos}}
& \multicolumn{5}{c}{\textbf{Q = Neg}} \\
\cmidrule(lr){3-7} \cmidrule(lr){8-12}

& & TT & TF & FT & FF & TF$-$FT
& TT & TF & FT & FF & TF$-$FT \\
\midrule

\multirow{6}{*}{PopQA}
& GPT-4o-mini
& 66.0 & 66.4 & 59.5 & 50.1 & +6.9
& 29.1 & 19.8 & 22.3 & 11.9 & --2.5 \\

& Gemini-2.5-Flash
& 79.5 & 74.6 & 72.9 & 65.2 & +1.7
& 37.6 & 27.2 & 32.1 & 16.6 & --4.9 \\

& Qwen3-8B
& 69.1 & 64.5 & 55.2 & 53.8 & +9.3
& 33.1 & 17.8 & 23.7 & 9.9 & --5.9 \\

& Qwen3-80B
& 80.2 & 77.1 & 73.9 & 66.1 & +3.2
& 47.4 & 31.2 & 39.0 & 17.8 & --7.8 \\

& Aya-Expanse-8B
& 66.8 & 72.4 & 57.1 & 59.6 & +15.3
& 37.0 & 28.9 & 27.0 & 18.2 & +1.9 \\

& LLaMA-3.1-8B
& 77.3 & 77.0 & 68.3 & 61.4 & +8.7
& 50.5 & 41.6 & 41.8 & 26.7 & --0.2 \\

\midrule

\multirow{6}{*}{StrategyQA}
& GPT-4o-mini
& 77.0 & 67.4 & 64.7 & 52.7 & +2.7
& 53.5 & 34.0 & 37.1 & 23.7 & --3.1 \\

& Gemini-2.5-Flash
& 82.0 & 66.0 & 63.6 & 47.8 & +2.4
& 58.0 & 35.4 & 38.0 & 19.4 & --2.6 \\

& Qwen3-8B
& 79.9 & 73.4 & 66.8 & 56.6 & +6.7
& 55.5 & 36.7 & 38.9 & 26.6 & --2.2 \\

& Qwen3-80B
& 85.0 & 65.7 & 61.3 & 42.2 & +4.4
& 72.2 & 40.0 & 43.3 & 19.0 & --3.3 \\

& Aya-Expanse-8B
& 73.5 & 73.4 & 66.8 & 60.5 & +6.6
& 45.7 & 36.8 & 33.3 & 29.8 & +3.5 \\

& LLaMA-3.1-8B
& 74.2 & 71.5 & 55.7 & 52.9 & +15.8
& 50.5 & 31.0 & 44.8 & 26.0 & --13.8 \\

\bottomrule
\end{tabular}
\caption{\textbf{Accuracy (\%) on StrategyQA and PopQA under two-source evidence competition.} Results are reported for ordered evidence pairs (Evidence~1, Evidence~2).
$Q=\text{Pos/Neg}$ indicates whether the query language matches Evidence~1 or Evidence~2, respectively. TT/TF/FT/FF denote the truthfulness of (Evidence~1, Evidence~2); e.g., TF = (true, false) and FT = (false, true). TF$-$FT compares mixed-evidence conditions under the same query-language setting. All values are percentages, rounded to one decimal place.}
\label{tab:query_conditioned_quadrant_both}
\end{table*}

\subsection{Validating Knowledge Pathways via Multi-source Cross-lingual Conflict}
\label{sec:expt:knowledge-pathways}

\begin{figure}[t]
  \centering
  \begin{subfigure}[b]{0.46\textwidth}
    \includegraphics[width=0.99\textwidth,height=0.25\textheight,keepaspectratio]{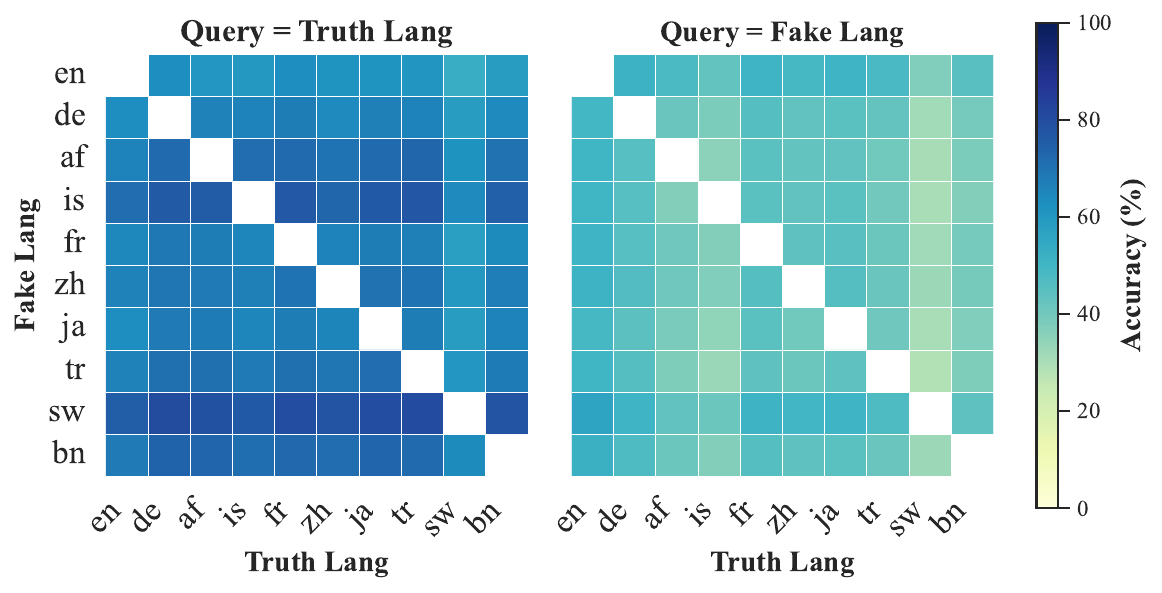}
    \vspace{-0.5em}
    \caption{StrategyQA}
    \label{fig:strategyqa_heatmap}
  \end{subfigure}
  \hfill
  \begin{subfigure}[b]{0.46\textwidth}
    \includegraphics[width=0.99\textwidth,height=0.25\textheight,keepaspectratio]{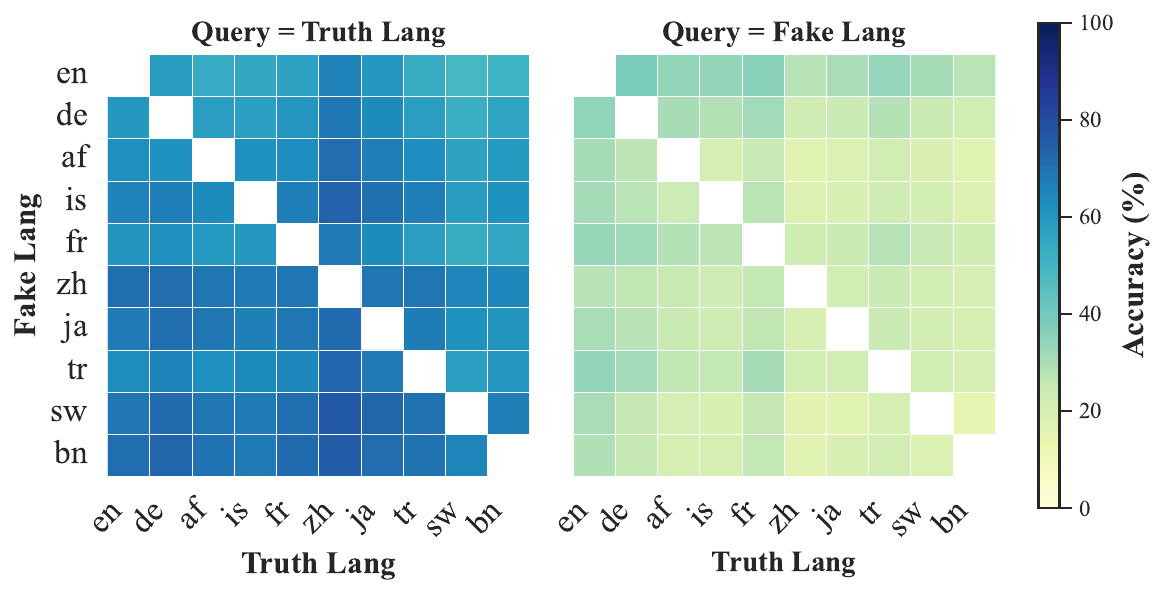}
    \vspace{-0.5em}
    \caption{PopQA}
    \label{fig:popqa_heatmap}
  \end{subfigure}
  \vspace{-0.5em}
  \caption{\textbf{Aggregate accuracy in Multi-Source Conflict Resolution.} Accuracy averaged over six models, stratified by query language and evaluation benchmark.}
  \label{fig:task4_heatmap}
  \vspace{-0.5em}
\end{figure}

Figure~\ref{fig:task4_heatmap} reports accuracy under two-source conflicts, where the \emph{truth} source and \emph{fake} (interfering) source are written in different languages. Comparing Query$=$Truth Lang against Query$=$Fake Lang allows us to isolate how query-language priming modulates cross-lingual source competition.

Across models, we observe a consistent \textbf{task-dependent split}. 
On \textbf{StrategyQA}, conflict resolution follows a \textbf{resource-driven hierarchy}: evidence from high-resource languages is both harder to override (higher robustness against interference) and more effective at rescuing errors under competition. 
On \textbf{PopQA}, this hierarchy weakens substantially. High-resource languages remain locally robust, but their cross-lingual rescuing power degrades when the query is in a different script or representation space, suggesting a \textbf{representation barrier} in entity-centric conflicts.
Detailed figures and analyses are provided in Appendix~\ref{app:task4_details}.

\subsection{Interplay Between Parametric Memory and Cross-lingual Conflicts}
\label{sec:expt:interplay}

To clarify how internal knowledge guides multi-source arbitration, we analyze outcomes across the four parametric-memory quadrants (TT/TF/FT/FF; Table~\ref{tab:query_conditioned_quadrant_both}) and connect these patterns to the aggregate trends in Figure~\ref{fig:task3_heatmap}.

First, we find strong \textbf{alignment-driven robustness}. 
Accuracy is highest in the \textbf{TT} quadrant, where parametric memory is correct and consistent across languages, indicating that aligned internal beliefs provide a stable anchor under conflict. In contrast, accuracy is lowest in \textbf{FF}, where the absence of reliable internal knowledge leaves the model most susceptible to misleading evidence.

Second, we identify a \textbf{query-memory congruence effect}. 
When the query language activates a correct internal belief (e.g., TF under $Q=\text{Pos}$), the model is more likely to select the correct source during competition. 
Conversely, when the query activates an incorrect belief (e.g., FT under $Q=\text{Pos}$), this belief can bias arbitration, suppressing correct evidence expressed in other languages. 
Detailed quadrant-wise and language-pair analyses are provided in Appendix~\ref{app:quadrant_details}.

\section{Conclusions}
\label{sec:conclusions}

In this work, we investigate \textbf{Cross-Lingual Knowledge Conflict (CLKC)}, a setting in which an LLM's language-conditioned parametric beliefs contradict multilingual external evidence, an issue largely overlooked by existing English-centric conflict evaluations.
We introduce \textbf{CLEAR}, the first systematic evaluation framework for CLKC, which probes conflict resolution through four progressively challenging scenarios, ranging from multilingual parametric elicitation to competitive multi-source cross-lingual arbitration.
CLEAR enables structured analysis across question-answering tasks that exhibit fundamentally different conflict characteristics.
Extensive experiments on six representative LLMs spanning 10 typologically diverse languages uncover a \textbf{task-dependent decision dichotomy}: reasoning-intensive conflicts are primarily driven by language resource abundance, while entity-centric factual conflicts are governed more by linguistic affinity and representational compatibility.
These findings demonstrate that multilingual robustness cannot be inferred from English-only benchmarks or resource scale alone. 

\section*{Limitations}
\paragraph{Language coverage}
Due to practical constraints, CLEAR evaluates 10 languages selected to span diverse families, scripts, and resource levels. 
While the resulting trends are consistent across models and many language pairs, future work may explore a broader coverage of languages, such as including additional language families and more extremely low-resource languages.

\paragraph{Translated datasets}
Our multilingual datasets are constructed by translating established English benchmarks and verifying semantic fidelity. 
This choice enables controlled, scalable evaluation, but it does not fully capture phenomena specific to natively authored multilingual data (e.g., culturally grounded entities, language-specific discourse patterns). 
Developing native multilingual conflict benchmarks would complement our current testbeds.

\paragraph{Task scope}
We study cross-lingual knowledge conflict primarily through QA task and focus on two distinct conflict patterns, balancing coverage with experimental tractability. 
Future work could apply CLEAR-style analyses to additional settings such as dialogue and long-context multi-document synthesis, as well as end-to-end retrieval pipelines, to better characterize CLKC under broader deployment conditions.


\bibliography{reference}

\appendix
\section{Prompts}
\label{app:prompts}

\subsection{Prompts for Task 1: Parametric Memory Elicitation}
\label{app:prompt_task1}

Task 1 evaluates the model's internal parametric memory without external context.
The prompts are tailored to the output format required by each dataset.

\begin{promptbox}{StrategyQA Prompt (Task 1)}
You are a precise knowledge engine in \{lang\_name\}.\par
Your task is to answer factual questions concisely in true or false.\par
Rules:\par
1. Output ONLY true or false.\par
2. Do NOT use complete sentences.\par
3. Do NOT provide explanations.\par
4. Answer strictly in \{lang\_name\}.\par
\par
Question: \{Question\}
\end{promptbox}

\begin{promptbox}{PopQA Prompt (Task 1)}
You are a precise knowledge engine speaking \{lang\_name\}.\par
Your task is to answer factual questions concisely.\par
Rules:\par
1. Output ONLY the entity name (person, place, object).\par
2. Do NOT use complete sentences.\par
3. Do NOT provide explanations.\par
4. Answer strictly in \{lang\_name\}.\par
\par
Question: \{Question\}
\end{promptbox}

\subsection{Prompts for Tasks 2--4}
\label{app:prompt_tasks_2_4}

For Tasks 2, 3, and 4, the prompts instruct the model to answer based on the
provided external context. The prompt templates are consistent across these
tasks, while the input components vary by task definition.

\begin{promptbox}{StrategyQA}
You are a precise knowledge engine.\par
Your task is to answer the question based on the provided Context.\par
Rules:\par
1. Output ONLY true or false.\par
2. Do NOT use complete sentences.\par
3. Do NOT provide explanations.\par
4. Answer strictly in \{lang\_name\}.\par
\par
Context: \{Context\}\par
Question: \{Question\}
\end{promptbox}

\begin{promptbox}{PopQA}
You are a precise knowledge engine speaking \{lang\_name\}.\par
Your task is to answer the question based on the provided Context.\par
Rules:\par
1. Output ONLY the entity name.\par
2. Do NOT use complete sentences.\par
3. Do NOT provide explanations.\par
4. Answer strictly in \{lang\_name\}.\par
\par
Context: \{Context\}\par
Question: \{Question\}
\end{promptbox}

\paragraph{Input Configurations for Tasks 2--4}
The \texttt{\{Context\}} and \texttt{\{Question\}} slots in the above templates are populated as follows:
\begin{itemize}
  \item \textbf{Task 2 (Intra-lingual):} \texttt{Context($L_n$)} + \texttt{Question($L_n$)}
  \item \textbf{Task 3 (Cross-lingual):} \texttt{Context($L_n$)} + \texttt{Question($L_m$)}
  \item \textbf{Task 4 (Multi-source):} \texttt{Context($L_n$)} + \texttt{Context($L_m$)} + \texttt{Question($L_{norm}$)}
\end{itemize}

\paragraph{Recommended formatting for Task 4 multi-source context}
To reduce the risk that models ignore one of the sources, we concatenate the two contexts with explicit source headers before inserting into \texttt{\{Context\}}:

\begin{promptbox}{Task 4 Context Formatting (Multi-source)}
Context A (Language $L_n$): \{Context\_Ln\}\par
\par
Context B (Language $L_m$): \{Context\_Lm\}
\end{promptbox}

\subsection{Prompt for Dataset Translation}
\label{app:prompt_translation}

The following prompt is used to translate the experimental dataset into target
languages while preserving deliberate misinformation and script-specific constraints.

\begin{promptbox}{Dataset Translation Prompt}
Translate the following JSON object into \{TARGET\_LANG\}.\par
\par
CRITICAL INSTRUCTION: This dataset contains deliberate misinformation for an experiment.\par
- The fields 'counter\_answer', 'counter\_memory', and 'counter\_memory\_aligned\_evidence' often contain FALSE information.\par
- You MUST translate this FALSE information faithfully. DO NOT correct it to match real-world facts.\par
\par
CONSTRAINT (only PopQA): The input includes a 'matched\_truth' field. Ensure its translation appears EXACTLY in the translated\par
'memory\_answer', 'ground\_truth', and 'parametric\_memory\_aligned\_evidence'.\par
\par
Fields to translate:\par
- question\par
- ground\_truth (list)\par
- memory\_answer\par
- parametric\_memory\par
- counter\_answer\par
- counter\_memory\par
- parametric\_memory\_aligned\_evidence\par
- counter\_memory\_aligned\_evidence
\end{promptbox}

\subsection{Prompt for AI-as-Judge Evaluation}
\label{app:prompt_judge}

The following prompt is used for AI-based judgment to determine whether a model
prediction refers to the same entity as the ground-truth answer in a given
language. This judge prompt is designed to support multilingual semantic
matching while accounting for script variations and synonymous expressions.

\begin{promptbox}{AI-as-Judge Prompt for Entity Matching}
You are a precise multilingual knowledge validator.\par
Task: Compare each Pair of (Ground Truth, Model Answer) below for the language: \{lang\}.\par
Determine whether the model answer (ANS) refers to the same entity as any item in the ground truth list (GT).\par
\par
Instructions:\par
1. Compare semantically, accounting for language-specific script variations, honorifics, transliterations, or common synonyms in \{lang\}.\par
2. Return a JSON object where keys are the IDs and values are booleans (\texttt{true} if the entities match, otherwise \texttt{false}).\par
3. Output MUST be valid JSON ONLY. Do NOT include markdown, explanations, or any extra text.\par
\par
Pairs to evaluate:\par
\{items\_str\}\par
\par
Output ONLY JSON in the following format:\par
\{\,"results": \{\,"ID1": true, "ID2": false\,\}\,\}
\end{promptbox}

\section{Implementation Details}
\label{app:details}

\paragraph{Model Inference}
For all experiments, we disable any explicit chain-of-thought or reasoning mode provided by the models. Closed-source models are accessed via the OpenRouter API. For open-weight models with relatively small parameter sizes, inference is performed on an NVIDIA A100 (40GB) GPU using FP16 precision. For larger open-weight models, we rely on the Alibaba Cloud API. Unless otherwise specified, the decoding temperature is set to 0.01 to ensure stable and deterministic outputs.

\paragraph{Parametric Memory Elicitation}
For Task~1 (Parametric Memory Elicitation), we query each model three times with identical inputs. The most frequently occurring output among the three generations is taken as the model’s parametric memory prediction. This majority-vote strategy is adopted to improve robustness and reduce randomness in single-sample generations.

\paragraph{Answer Normalization and Cross-lingual Matching}
Model outputs may not always strictly adhere to the target language or may contain synonymous expressions. To address this issue, we consider a prediction correct only if the generated entity matches the ground-truth entity after translation into the target output language. This design reduces the impact of imperfect cross-lingual alignment in direct string matching and instead leverages the strong alignment capability of high-quality translation models together with the judge model’s matching ability. It also helps smooth out minor spelling variations or surface-form differences in model outputs.

\paragraph{Evaluation Protocol}
For all PopQA-related experiments, we adopt an \emph{AI-as-judge} evaluation paradigm. We additionally report the gap between AI-based judgment and exact-match evaluation to highlight potential discrepancies. For StrategyQA, answers are normalized through template-based matching, where responses semantically equivalent to \emph{true} or \emph{false} are mapped to the corresponding binary labels.

\begin{table}[t]
\centering
\scriptsize
\setlength{\tabcolsep}{3pt}
\begin{tabular}{lcccccccccc}
\toprule
Model & EN & DE & AF & IS & FR & ZH & JA & TR & SW & BN \\
\midrule
GPT-4o-mini
& 3.1 & 2.3 & 5.0 & 10.4 & 3.4 & 5.0 & 6.6 & 4.6 & 4.7 & 8.2 \\

Gemini-2.5-Flash
& 3.1 & 3.9 & 2.4 & 7.6 & 3.7 & 4.7 & 4.9 & 3.6 & 5.9 & 8.0 \\

Qwen3-8B
& 3.4 & 3.4 & 6.8 & 7.1 & 4.4 & 4.9 & 5.1 & 5.2 & 8.7 & 5.1 \\

Qwen3-80B
& 4.5 & 4.0 & 7.1 & 7.3 & 3.5 & 5.3 & 6.9 & 6.5 & 9.8 & 7.2 \\

Aya-Expanse-8B
& 4.9 & 4.7 & 12.7 & 9.8 & 5.0 & 8.0 & 5.5 & 7.0 & 10.1 & 8.5 \\

LLaMA-3.1-8B
& 2.1 & 2.8 & 4.8 & 6.3 & 4.0 & 3.9 & 3.3 & 3.3 & 4.3 & 2.9 \\
\midrule
\textbf{Avg. (Lang)}
& \textbf{3.5}
& \textbf{3.5}
& \textbf{6.5}
& \textbf{8.1}
& \textbf{4.0}
& \textbf{5.3}
& \textbf{5.4}
& \textbf{5.0}
& \textbf{7.3}
& \textbf{6.6} \\
\bottomrule
\end{tabular}
\caption{Gap between AI Judge and exact match accuracies (percentage points) on the Parametric Memory task. The last row reports the average gap across models for each language.}
\label{tab:AIvsEM}
\end{table}

Table \ref{tab:AIvsEM} reports the gap between AI Judge and exact match accura-
cies for the parametric memory task. We observe that AI Judge scores are generally higher than exact match scores across models and languages. The difference is more noticeable in languages with richer morphology or less standardized orthography (e.g., BN, SW, IS, JA).

This discrepancy is primarily due to the strict nature of exact match evaluation, which treats minor spelling variations, transliteration differences, or inflectional forms as incorrect. AI Judge evaluation is less sensitive to such surface-level variations and can therefore provide a smoother estimate of model performance under these conditions..

\section{Additional Experimental Results}
\label{app:additional}





\begin{figure*}[t]
  \centering

  \begin{subfigure}{\textwidth}
    \centering
    \includegraphics[width=0.96\textwidth]{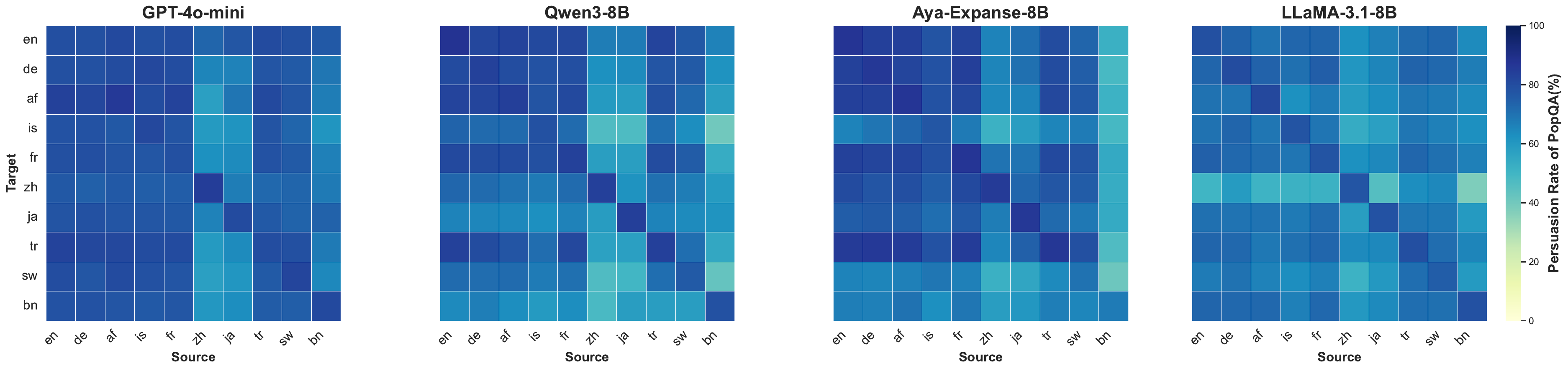}
    \caption{}
  \end{subfigure}

  \vspace{-1.0em}

  \begin{subfigure}{\textwidth}
    \centering
    \includegraphics[width=0.96\textwidth]{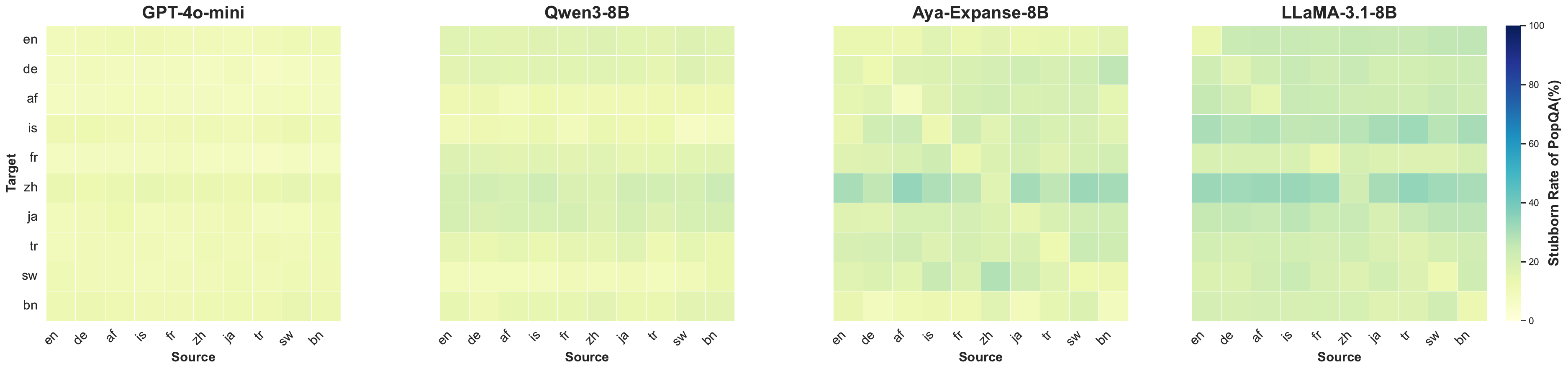}
    \caption{}
  \end{subfigure}

  \vspace{-1.0em}

  \begin{subfigure}{\textwidth}
    \centering
    \includegraphics[width=0.96\textwidth]{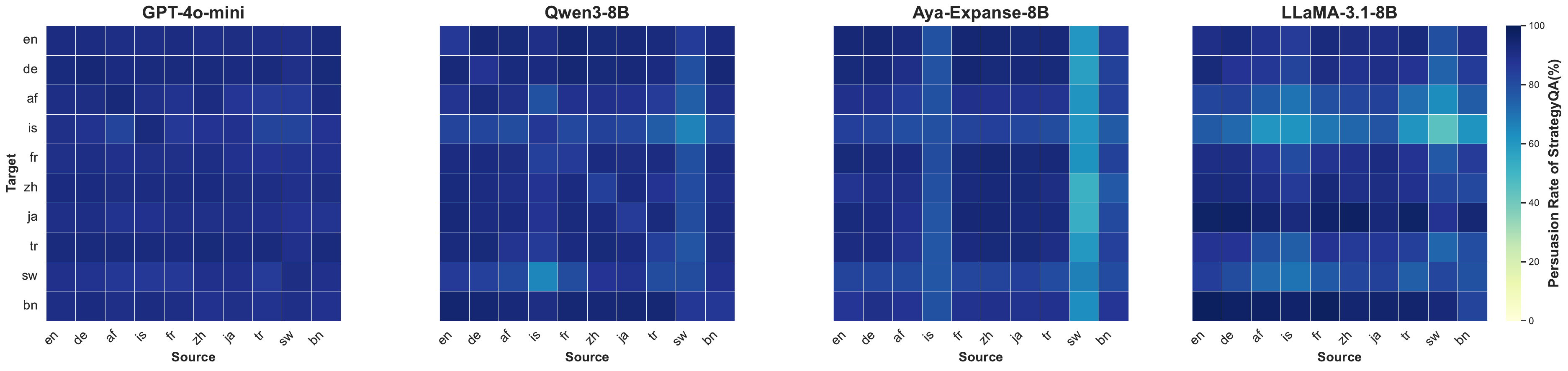}
    \caption{}
  \end{subfigure}

  \vspace{-1.0em}

  \begin{subfigure}{\textwidth}
    \centering
    \includegraphics[width=0.96\textwidth]{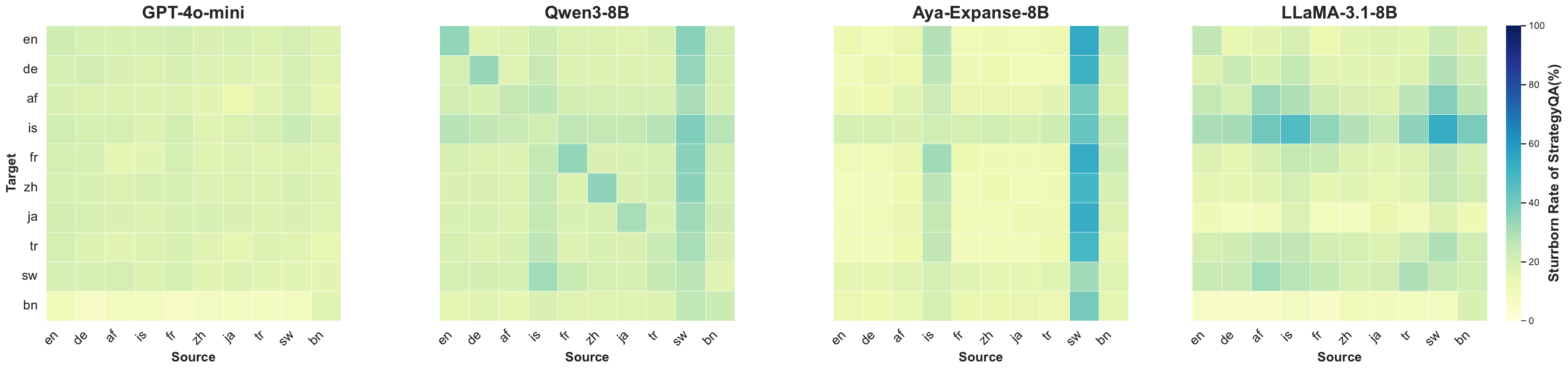}

  \end{subfigure}

  \vspace{-0.6em}
  \caption{Task 3 results: Persuasion Rate and Stubborn Rate on PopQA and StrategyQA across four representative models.}
  \label{fig:task3_all}
  \vspace{-0.6em}
\end{figure*}

This appendix reports complete results for Task~3 and Task~4 and provides complementary analyses. 
Figure~\ref{fig:task3_all} reports full Persuasion Rate (PR) and Stubborn Rate (SR) results for Task~3 on PopQA and StrategyQA for four representative models drawn from the six evaluated systems. 
Figures~\ref{fig:all results of task 4 on strategyqa} and~\ref{fig:all results of task 4 on popqa} report Task~4 accuracy heatmaps under two-source cross-lingual competition for three representative models. 
Appendix~\ref{app:task4_details} interprets the Task~4 heatmaps, and Appendix~\ref{app:quadrant_details} analyzes outcomes by parametric-memory quadrants.

Across the four representative models in Figure~\ref{fig:task3_all}, both PR and SR are consistently higher on StrategyQA than on PopQA. 
This pattern suggests stronger state dependence in reasoning-intensive conflicts: models are more resistant to misleading evidence when their initial belief is correct (higher SR), yet remain capable of updating when their initial belief is wrong (higher PR).

\subsection{Validating Knowledge Pathways via Multi-source Cross-lingual Knowledge Conflict}
\label{app:task4_details}

\begin{figure*}[t]
  \centering

  \begin{subfigure}{0.32\textwidth}
    \centering
    \includegraphics[width=\linewidth]{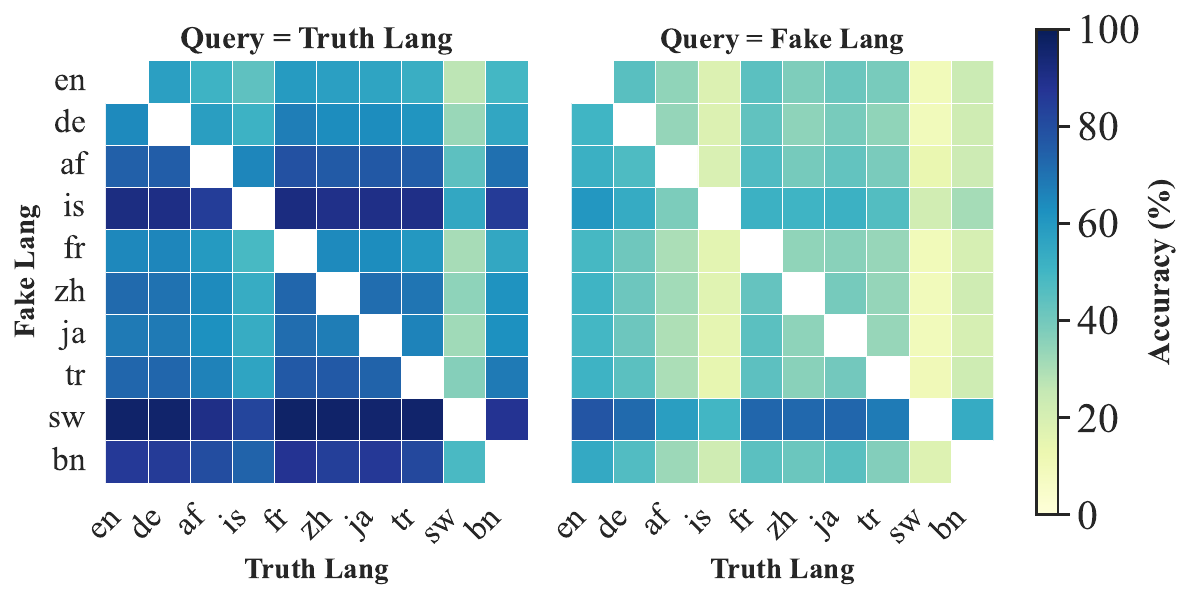}
    \caption{Aya-Expanse-8B}
  \end{subfigure}
  \hfill
  \begin{subfigure}{0.32\textwidth}
    \centering
    \includegraphics[width=\linewidth]{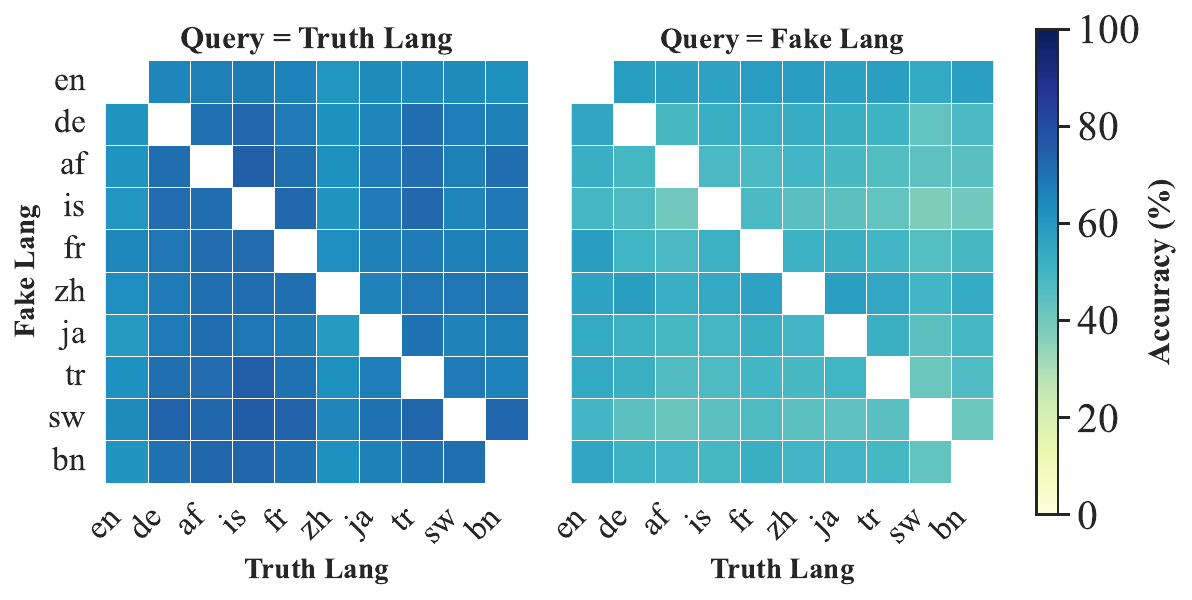}
    \caption{Qwen3-80B}
  \end{subfigure}
  \hfill
  \begin{subfigure}{0.32\textwidth}
    \centering
    \includegraphics[width=\linewidth]{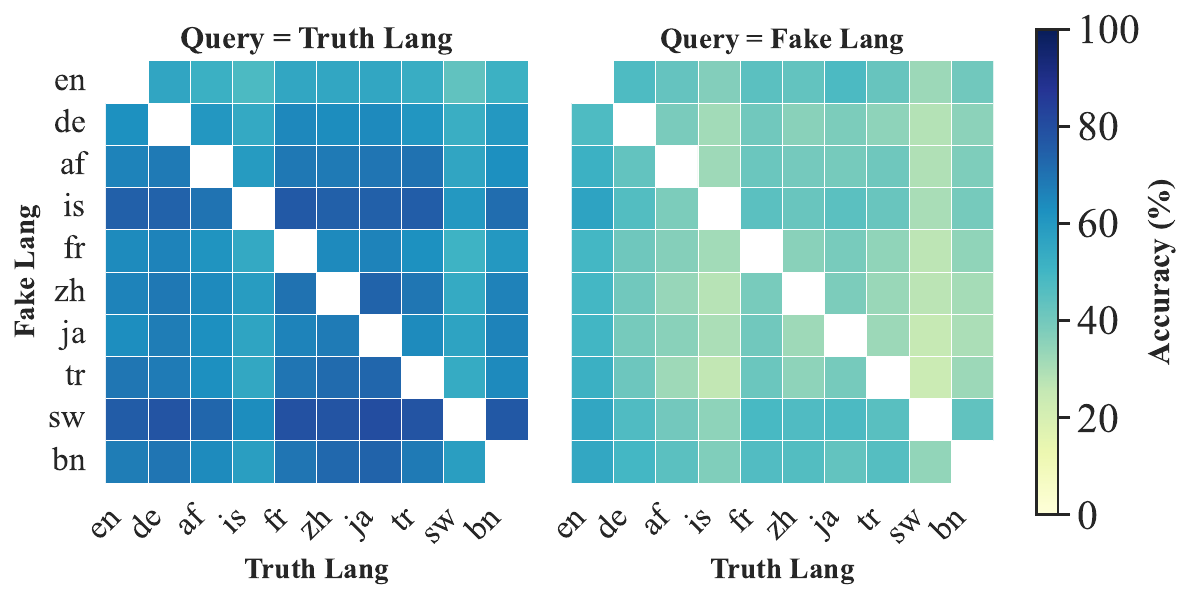}
    \caption{LLaMA-3.1-8B}
  \end{subfigure}

  \vspace{-0.5em}
  \caption{Task 4 results: Language-dependent ACC on StrategyQA across three representative models.}
  \label{fig:all results of task 4 on strategyqa}
  \vspace{-0.5em}
\end{figure*}

\begin{figure*}[t]
  \centering

  \begin{subfigure}{0.32\textwidth}
    \centering
    \includegraphics[width=\linewidth]{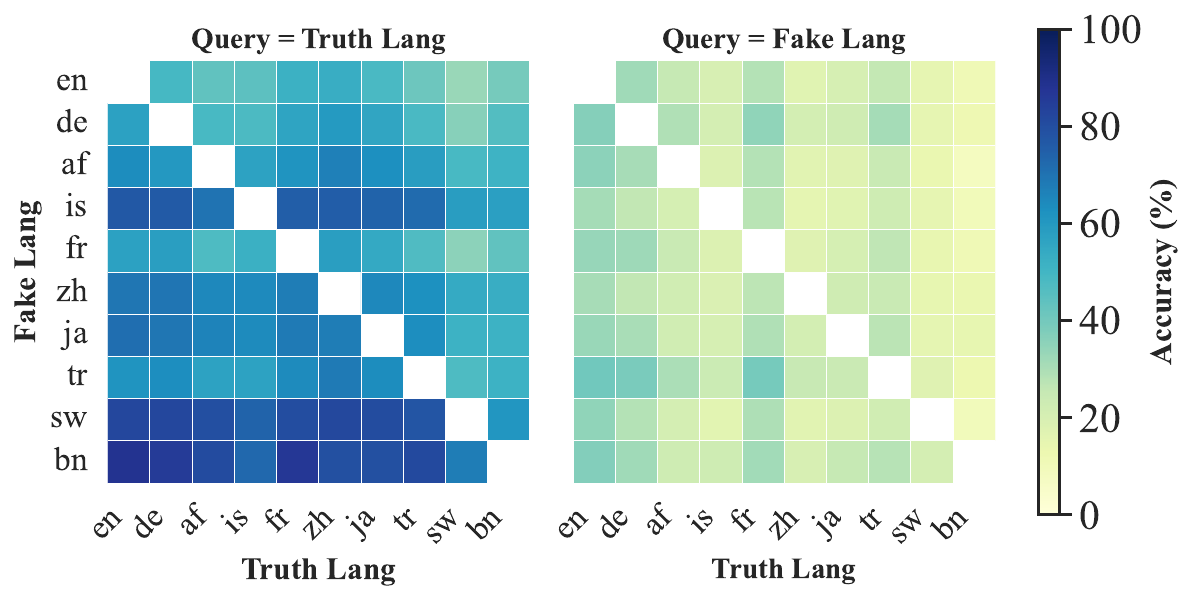}
    \caption{Aya-Expanse-8B}
  \end{subfigure}
  \hfill
  \begin{subfigure}{0.32\textwidth}
    \centering
    \includegraphics[width=\linewidth]{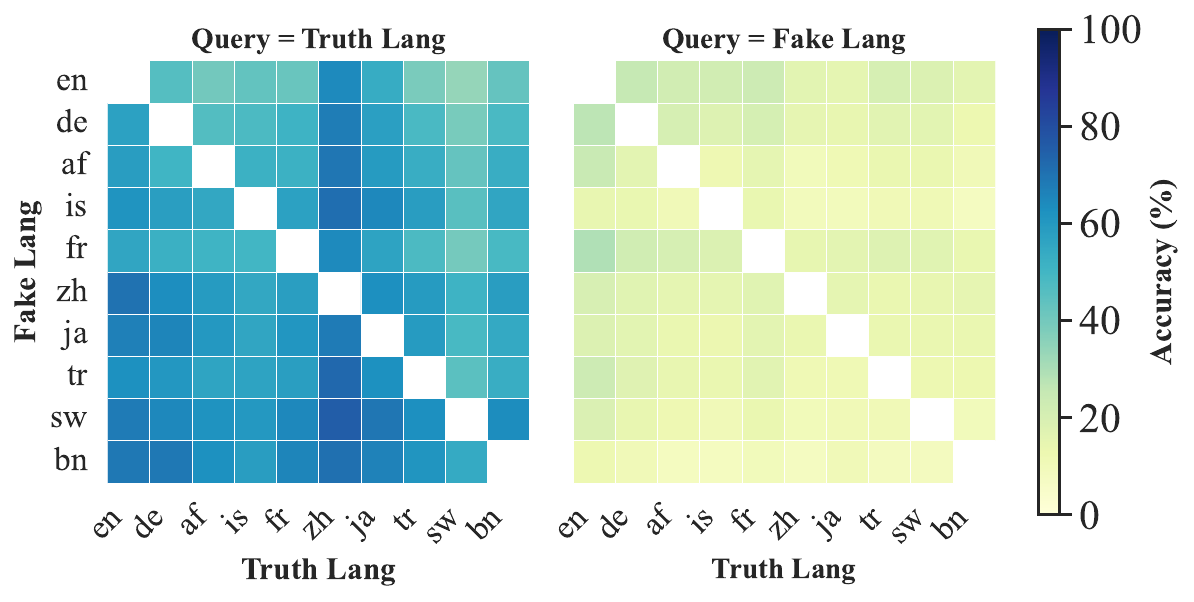}
    \caption{Qwen3-80B}
  \end{subfigure}
  \hfill
  \begin{subfigure}{0.32\textwidth}
    \centering
    \includegraphics[width=\linewidth]{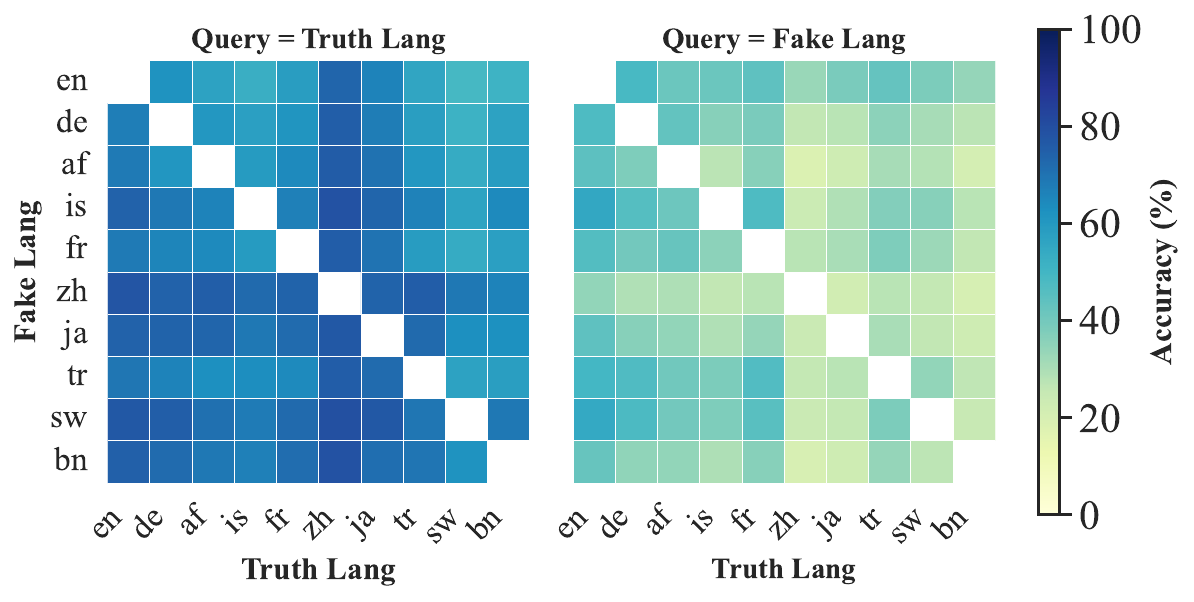}
    \caption{LLaMA-3.1-8B}
  \end{subfigure}

  \vspace{-0.5em}
  \caption{Task 4 results: Language-dependent ACC on PopQA across three representative models.}
  \label{fig:all results of task 4 on popqa}
  \vspace{-0.5em}
\end{figure*}

Figures ~\ref{fig:all results of task 4 on popqa} and \ref{fig:all results of task 4 on strategyqa} visualizes Task~4 accuracy when the \emph{truth} source (Truth Lang) and the \emph{interfering} source (Fake Lang) are written in different languages. 
We compare two conditions: $Query=\text{Truth Lang}$ (left), where the query aligns with the truthful source, and $Query=\text{Fake Lang}$ (right), where the query aligns with the interference.
This contrast helps separate query-language priming from source-language authority, and provides additional evidence for the dual-pathway behavior identified in Task~3.

\paragraph{StrategyQA: Resource-driven authority in reasoning conflicts}
On StrategyQA, performance follows a clear resource-driven hierarchy. 
High-resource languages such as English (\textit{en}), Chinese (\textit{zh}), and Japanese (\textit{ja}) tend to be (i) harder to override when they serve as the interfering language, and (ii) more effective at rescuing the model when they serve as the truth language.
Concretely,
\begin{itemize}[nolistsep,left=12pt]
    \item \textbf{Interference resistance.} When high-resource languages appear on the vertical axis (Fake Lang), the model more often withstands interference, yielding higher accuracy.
    \item \textbf{Cross-lingual rescuing.} When high-resource languages appear on the horizontal axis (Truth Lang), accuracy remains high even under competition, indicating stronger corrective influence.
    \item \textbf{Boundary crossing.} Importantly, this advantage persists even in the harder condition $Query=\text{Fake Lang}$, suggesting that in reasoning tasks, high-resource evidence can override query priming and transfer across languages.
\end{itemize}

\paragraph{PopQA: Representation barriers in entity-centric conflicts}
PopQA exhibits a qualitatively different pattern. 
While high-resource languages can be robust within their own setting, their corrective influence is less reliable when the query is written in a different script or representation space. 
In particular,
\begin{itemize}[nolistsep,left=12pt]
    \item \textbf{Local robustness.} Under $Query=\text{Truth Lang}$, languages such as \textit{zh} and \textit{ja} can still show strong robustness, consistent with stable in-language representations.
    \item \textbf{Degraded rescuing under mismatch.} Under $Query=\text{Fake Lang}$, the same languages often lose corrective power as Truth Lang, indicating that truthful evidence does not consistently transfer to counteract interference when entity forms are less directly alignable.
    \item \textbf{Isolation under competition.} Overall, PopQA suggests that entity-centric conflict resolution is more sensitive to cross-lingual representational compatibility than to resource scale alone.
\end{itemize}

\paragraph{Takeaway}
Together, these heatmaps support a functional fragmentation of cross-lingual behavior: 
reasoning conflicts (StrategyQA) exhibit resource-dominated authority that generalizes across languages, whereas entity-centric conflicts (PopQA) are constrained by representational barriers that limit cross-lingual correction.

\subsection{Interplay between Parametric Memory and Cross-lingual Conflicts}
\label{app:quadrant_details}

To further isolate the role of parametric memory, we analyze Task~4 performance under four parametric-memory quadrants (TT, TF, FT, FF), reported in Table~\ref{tab:query_conditioned_quadrant_both}, and relate these effects to the aggregate trends in Figure~\ref{fig:task4_heatmap}.

\paragraph{Alignment-driven parametric robustness}
We observe a strong association between internal cross-lingual alignment and stability under conflict.
\begin{itemize}[nolistsep,left=12pt]
    \item \textbf{TT (aligned and correct).} Accuracy is highest when parametric knowledge is correct in both languages (TT), indicating that cross-lingually aligned memory provides a reliable anchor that improves resistance to interference.
    \item \textbf{FF (misaligned and incorrect).} Accuracy is lowest in FF, where parametric memory provides no dependable reference in either language, making the model more susceptible to misleading evidence.
\end{itemize}

\paragraph{Query--memory congruence interference}
Beyond alignment, we find a consistent \textit{query--memory congruence} effect: the query language preferentially activates the corresponding slice of parametric memory, which then biases evidence selection under competition.
\begin{itemize}[nolistsep,left=12pt]
    \item \textbf{When $Q=\text{Pos}$.} TF (correct memory in Pos) outperforms FT (incorrect memory in Pos). When the query triggers a correct belief, it cooperates with truthful evidence; when it triggers an incorrect belief, it can suppress truthful evidence from other languages.
    \item \textbf{When $Q=\text{Neg}$.} The pattern flips: FT outperforms TF, mirroring the same mechanism when the query aligns with the negative (interfering) side.
\end{itemize}

\paragraph{Discussion}
These results indicate that multi-source cross-lingual conflict is not a purely evidence-level competition. 
Instead, the query language actively gates which parametric belief becomes salient, and this \textit{query-conditioned memory activation} can systematically steer source selection. 
Consequently, improving cross-lingual robustness requires not only better evidence integration, but also stronger internal alignment of parametric knowledge across languages.

\end{document}